\documentclass[letterpaper, 10 pt, conference]{ieeeconf}  

\usepackage{amsmath,amsfonts}
\usepackage{algorithmic}
\usepackage{algorithm}
\usepackage{array}
\usepackage[caption=false,font=normalsize,labelfont=sf,textfont=sf]{subfig}
\usepackage{textcomp}
\usepackage{stfloats}
\usepackage{url}
\usepackage{verbatim}
\usepackage{graphicx}
\usepackage{multirow} 
\usepackage{multicol}
\usepackage[hidelinks]{hyperref}
\usepackage{cite}
\usepackage{tikz,xcolor}
\usepackage{bm}
\usepackage{fancyhdr}
\usepackage{booktabs}
\usepackage{lettrine}
\usepackage{verbatim}
\usepackage{threeparttable}
\usepackage{makecell}
\usepackage{xcolor}
\usepackage[T1]{fontenc}
\usepackage{cite}
\usepackage{balance}
\newcommand{\etal}{\textit{et al.}}

\IEEEoverridecommandlockouts                              

\title{\LARGE \bf
GaussianPU: A Hybrid 2D-3D Upsampling Framework for Enhancing Color Point Clouds via 3D Gaussian Splatting
}

\author{Zixuan Guo$^{1*}$, Yifan Xie$^{2*}$, Weijing Xie$^{3}$, Peng Huang$^{4}$, Fei Ma$^{\dagger}$$^{5}$  and Fei Richard Yu$^{5}$
\thanks{$^{*}$Both authors contributed equally to this work. $^{\dagger}$Corresponding author. (Corresponding email: mafei@gml.ac.cn)}
\thanks{$^{1}$Peking University $^{2}$Xi'an Jiaotong University $^{3}$Sun Yat-Sen University $^{4}$Nanjing University $^{5}$Guangdong Laboratory of Artificial Intelligence and Digital Economy (SZ)}%
}

\begin{document}

\maketitle
\thispagestyle{empty}
\pagestyle{empty}

\begin{abstract}
Dense colored point clouds enhance visual perception and are of significant value in various robotic applications.  However, existing learning-based point cloud upsampling methods are constrained by computational resources and batch processing strategies, which often require subdividing point clouds into smaller patches, leading to distortions that degrade perceptual quality. To address this challenge, we propose a novel 2D-3D hybrid colored point cloud upsampling framework (GaussianPU) based on 3D Gaussian Splatting (3DGS) for robotic perception.  This approach leverages 3DGS to bridge 3D point clouds with their 2D rendered images in robot vision systems. A dual scale rendered image restoration network transforms sparse point cloud renderings into dense representations, which are then input into 3DGS along with precise robot camera poses and interpolated sparse point clouds to reconstruct dense 3D point clouds.  We have made a series of enhancements to the vanilla 3DGS, enabling precise control over the number of points and significantly boosting the quality of the upsampled point cloud for robotic scene understanding.  Our framework supports processing entire point clouds on a single consumer-grade GPU, such as the NVIDIA GeForce RTX 3090, eliminating the need for segmentation and thus producing high-quality, dense colored point clouds with millions of points for robot navigation and manipulation tasks.  Extensive experimental results on generating million-level point cloud data validate the effectiveness of our method, substantially improving the quality of colored point clouds and demonstrating significant potential for applications involving large-scale point clouds in autonomous robotics and human-robot interaction scenarios.

\end{abstract}

\section{INTRODUCTION}
Colored point clouds augment traditional point clouds, which solely encompass geometric positional information, with the addition of color attributes. This enhancement considerably expands the data's dimensionality, providing a richer visual perception for robotic systems across various applications~\cite{diniz2021color,he2021towards} such as autonomous navigation, object manipulation, and human-robot interaction. The integration of color information enables robots to better interpret their environment, improving tasks like object recognition~\cite{oliva2007role}, scene understanding~\cite{cordts2016cityscapes}, and semantic mapping~\cite{kostavelis2015semantic}. However, the performance constraints of current robotic sensors often result in colored point clouds that are sparse and non-uniform, leading to an inevitable reduction in the robot's perception accuracy and decision-making capabilities. In response to this challenge, the application of upsampling techniques to increase the density of point clouds is of paramount importance in robotics. The aim is to generate denser, higher-quality colored point clouds that can significantly improve the robot's ability to interpret complex scenes, recognize objects with fine details, and interact more precisely with its surroundings. This enhanced perception capability is crucial for advancing robotic systems, enabling more sophisticated and reliable autonomous operations in applications ranging from industrial automation to search and rescue missions.

\begin{figure}
	\centering
	\includegraphics[width=0.9\columnwidth]{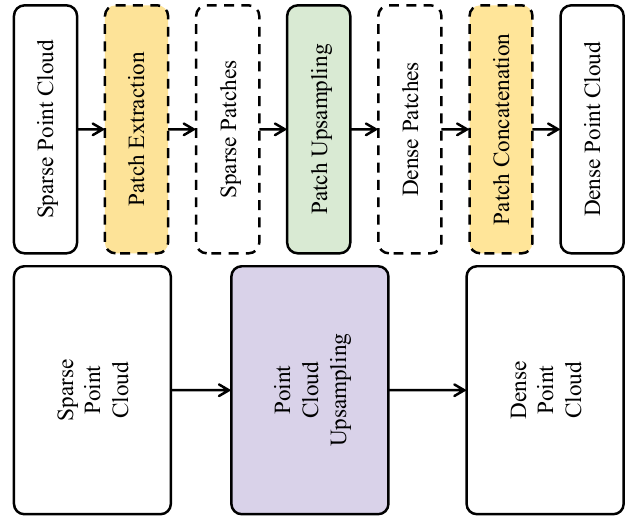} 
	\caption{The comparison between previous point cloud upsampling methods and ours. (Up) Previous methods primarily relied on patch-based upsampling, processing point clouds patch by patch. (Down) In contrast, our approach can directly upsample the entire point cloud, thereby avoiding the degradation in quality that can arise from the patch-based approach.
	}
	\label{figure1}
\end{figure}

Contemporary advanced point cloud upsampling techniques commonly employ deep learning architectures based on Multi-Layer Perceptrons (MLPs)~\cite{qi2017pointnet,wang2019dynamic,xie2023cross,xie2024hecpg}.
Despite the significant strides in performance achieved by these architectures, their considerable computational demands and the batch processing methodology employed in training often necessitate pre-processing of point clouds into patches. These patches are subsequently utilized as the basic input units for the model, a process that is exemplified in Figure \ref{figure1}. Upsampling models that segment point clouds into patches have the potential to create discontinuities between these segments. These inconsistencies can impair the overall quality and the perceptual experience of the point cloud. The issue is particularly relevant for color point clouds with a large quantity of points, which are intended for human visual perception.

In the realm of 3D robotic perception, neural rendering techniques based on point cloud data have seen significant advancements. 3D Gaussian Splatting (3DGS)~\cite{kerbl20233d} has emerged as a promising technology for robotic vision systems. 3DGS represents each point as a Gaussian distribution, blending these during rendering to form continuous surface representations. This approach considers position, density, and normal information, delivering accurate and realistic point cloud data for improved robot perception. 3DGS offers benefits in rendering quality, speed, and computational efficiency, making it ideal for real-time robotic applications. By leveraging parallel computing and optimizing parameters, 3DGS enables efficient processing of large-scale point clouds, enhancing a robot's ability to interpret complex environments and perform tasks such as navigation, object manipulation, and scene understanding with greater precision and responsiveness.

In this paper, we propose a novel point cloud upsampling framework for robotic perception, combining 3D Gaussian Splatting (3DGS) with a lightweight 2D image restoration network. Our approach addresses the challenges of processing large-scale point clouds in robotics applications. We introduce a dual-scale rendering restoration model to manage occlusion uncertainties in sparse point cloud renderings. Utilizing 3DGS as a bridge between 2D images and 3D point clouds, along with geometric and color priors from interpolated point clouds, we convert restored images into high-quality 3D point clouds. Our framework enables robots to generate large-scale colored point clouds efficiently on consumer-grade GPUs, enhancing their environmental understanding capabilities. Our contributions can be summarized as follows:
\begin{itemize}
\item We propose a novel color point cloud upsampling framework that combines 3DGS and  2D rendered image restoration models. This framework addresses the challenges of directly processing large-scale point clouds using traditional upsampling methods.
\item In order to deal with the front and rear occlusion problem of point clouds rendered images , we propose a dual scale rendering restoration method. A series of improvements tailored for adapting vanilla 3DGS to the point cloud upsampling task have effectively enhanced the quality of the upsampled point clouds and enabled precise control over the number of upsampled points.
\item Comprehensive experiments in robotic perception scenarios demonstrate that our proposed method significantly enhances the visual and spatial accuracy of sparse colored point clouds, improving a robot's ability to interpret and navigate complex environments.
\end{itemize}

\section{PRELATED WORK}

\subsection{Point Cloud Upsampling}
Early point cloud upsampling approaches primarily relied on optimization methods to upsample point clouds~\cite{alexa2003computing,huang2013edge}, which were heavily dependent on prior knowledge of the point cloud. With the advancement of deep learning, point cloud upsampling networks based on deep learning have emerged and achieved impressive results. PU-Net~\cite{yu2018pu} employs PointNet++~\cite{qi2017pointnet++} for hierarchical learning and multi-level feature aggregation to map the 3D coordinates of points to a feature space. MPU~\cite{yifan2019patch} employs an end-to-end, multi-step patch-based network to progressively upsample sparse 3D point sets. PU-GAN~\cite{li2019pu}introduces a GAN-based framework for point cloud upsampling that learns diverse point distributions and ensures uniformity across upsampled patches. PU-GCN~\cite{qian2021pu}, leveraging EdgeConv~\cite{wang2019dynamic} as its GCN backbone, introduces NodeShuffle for upsampling and Inception DenseGCN for multi-scale feature extraction, significantly advancing the point cloud upsampling pipeline. Grad-PU~\cite{he2023grad} introduces a novel framework for point cloud upsampling that can handle arbitrary upsampling rates by decomposing the problem into midpoint interpolation and refinement via point-to-point distance minimization. The above deep learning-based methods use multi-sampling MLP and EdgeConv as the basic modules of the model, which brings significant performance overhead and is difficult to directly process large-scale point clouds. 
Therefore, most existing solutions are based on less than 1,024 points. Point cloud patches are trained, which in turn makes it difficult to learn the relationship between patches. 
The point cloud upsampling method mentioned above mainly upsamples the geometry of colorless point clouds and does not utilize the color information characteristics of colored point clouds.

\subsection{Point-based Neural Rendering}
Aliev \etal~\cite{aliev2020neural} delivered an early pioneering work with Neural Point-Based Graphics, utilizing neural networks to enhance point cloud rendering, setting a precedent for subsequent innovations in image detail and texture from point data. 
With the proposal of Neural Radiance Fields (NeRF)~\cite{mildenhall2021nerf}, neural rendering has entered a stage of rapid development~\cite{tewari2022advances}. Yu \etal~\cite{xu2022point} further streamlined the rendering pipeline with a spatially coherent point-based multiplane image method,i.e. Point-NeRF improving the efficiency of synthesizing complex scenes. Recently, Schmitt~\etal introduced 3D Gaussian Splatting (3DGS)~\cite{kerbl20233d}, a real-time point cloud rendering technique that projects 3D points onto a 2D view plane using adaptive Gaussian kernels. By incorporating innovations such as sparse matrix Gaussian filtering and GPU optimization, 3DGS achieves interactive rendering frame rates while maintaining high quality, making it a promising approach for efficient point cloud visualization.

Currently, 3D Gaussian Splatting (3DGS)~\cite{kerbl20233d} has been successfully applied to various domains~\cite{chen2024survey}, such as dynamic 3D scene modeling~\cite{kocabas2023hugs,wu20234d,luiten2023dynamic,yang2023real}, artificial intelligence-generated content (AIGC)~\cite{chen2023text,tang2023dreamgaussian,li2023gaussiandiffusion,yi2023gaussiandreamer}, and autonomous driving~\cite{yan2024street,zhou2023drivinggaussian,xiong2024gauu}, achieving remarkable results. While significant progress has been made in the aforementioned areas, the advancements of 3DGS in low-level point cloud processing have been relatively limited. To the best of our knowledge, this paper is the first work that introduces the advantages of 3DGS to the task of colored point cloud upsampling.

\begin{figure*}[ht]  
\centering                  
\includegraphics[scale=0.55]{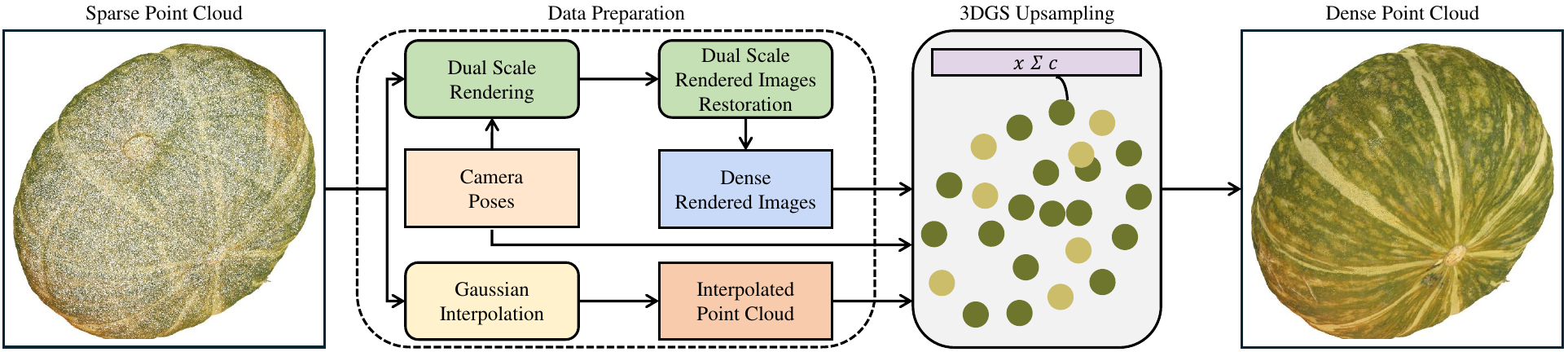} 
\caption{The framework of our proposed GaussianPU. In the data preparation stage, the sparse point cloud undergoes a rendering process followed by a restoration step to obtain dense rendered images. Simultaneously, the sparse point cloud is subjected to Gaussian interpolation to acquire a $R \times$ interpolated point cloud. The interpolated point cloud and dense rendered images obtained from the data preparation stage are fed into the 3DGS point cloud upsampling module along with the camera poses. Through iterative optimization, the 3DGS upsampling module generates a dense colored point cloud.}
\label{figure_pipeline}
\end{figure*}

\section{METHOD}
We present an upsampling pipeline for color point clouds as depicted in Figure \ref{figure_pipeline}. The entire upsampling framework can be divided into two modules: data preparation and 3DGS upsampling. The data preparation stage involves rendering the sparse point cloud, reconstructing the rendered images, and interpolating the sparse point cloud. Subsequently, the reconstructed rendered images and the interpolated point cloud, along with the camera poses, are fed into the upsampling module to obtain the upsampled dense point cloud.

\subsection{Preliminaries}
Consider a point cloud $P$ consisting of $N$ points, denoted as $P=\left\{p_{i} \in\mathbb{R}^6\right\}_{i=1}^{N}$, where each point $p_i$ is represented by a 6-dimensional vector. This vector encapsulates two fundamental properties: geometry and color. The geometric properties are characterized by the XYZ coordinates, which define the spatial position of each point within the 3D space, while the color attributes are represented by the RGB values associated with each point, providing the visual appearance information of the point cloud. Given a sparse point cloud $P_{sp}$ and an upsampling factor $R$, our objective is to generate a dense point cloud containing $R \times N$ points, represented as $ P_{ds} = \left\{ p_{i} \in \mathbb{R}^{6} \right\}_{i=1}^{R\times N} $. Moreover, we aim for $ P_{ds} $ to exhibit higher perceptual and geometric quality.

\subsection{Point Cloud Rendering and Rendered Image Restoration}
\emph{Point Cloud Rendering.}
In this work, we leverage Open3D~\cite{zhou2018open3d} to facilitate the rendering of point clouds. We encapsulate the rendering process through the following mathematical formulation: 
\begin{equation}
    I=Render(P,PS,IR,T),
\end{equation}
where $PS$ represents the point size, IR denotes the resolution of the rendered image, and K is the camera's extrinsic parameters, which are expressed by the following formula:
\begin{equation}
K =
\begin{bmatrix}
f_x & 1 & z_x \\
0 & f_y & z_y \\
0 & 0 & 1
\end{bmatrix},
\end{equation}
where $f_x$ and $f_y$ represent the camera's focal lengths in pixel units along the horizontal and vertical axes of the image sensor, respectively. The terms $z_x$ and $z_y$ denote the coordinates of the principal point,
The symbol $T\in\mathbb{R}^{4 \times 4}$ denotes the camera extrinsic parameters.

\begin{figure}[t]
    \begin{minipage}[t]{\columnwidth}
    \setlength{\tabcolsep}{1pt} 
    \centering
    \begin{tabular}{@{}cccc@{}}
    
        \hspace{-12pt}\includegraphics[width=0.33\textwidth]{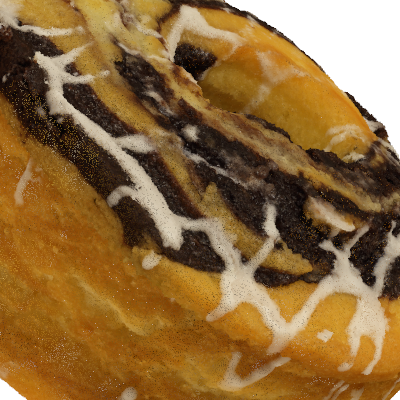} &
        \includegraphics[width=0.33\textwidth]{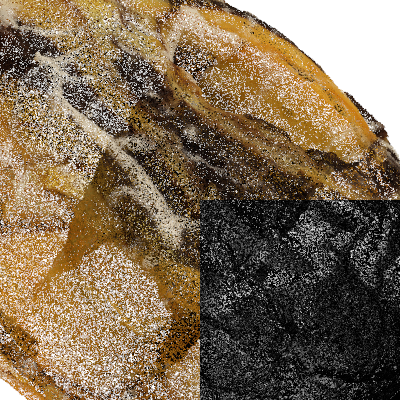} &
        \includegraphics[width=0.33\textwidth]{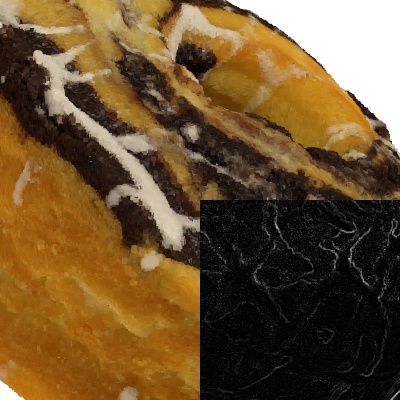} &\\
        \hspace{-12pt}\includegraphics[width=0.33\textwidth]{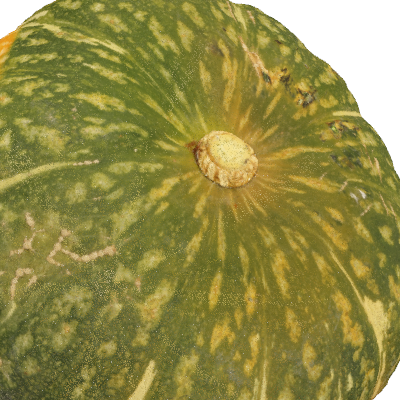} &
        \includegraphics[width=0.33\textwidth]{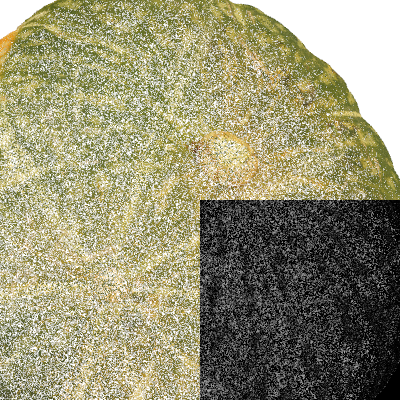} &
        \includegraphics[width=0.33\textwidth]{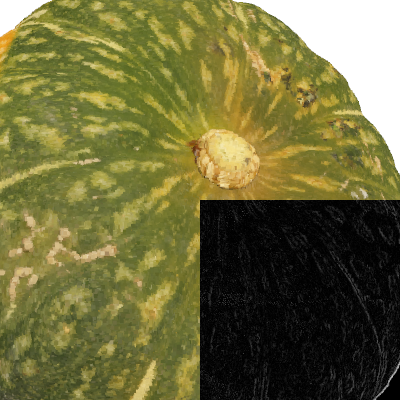} & \\
        \hspace{-12pt}{\footnotesize Ground-Truth} &
        {\footnotesize 1Pix} & 
        {\footnotesize 4Pix} & 
    \end{tabular}
        \caption{Sparse point cloud rendering with different point sizes and their errors compared to dense point cloud rendering.}
    \label{multi_size_renders}
    \end{minipage}
\end{figure}

In the setting of the point size $PS$, we have discovered that both large and small sizes of point renderings come with distinct advantages and disadvantages. As illustrated in Figure \ref{multi_size_renders}, when the $PS$ is set to 1, it retains more of the original information from the point cloud. However, this setting also allows noise points from the back of the point cloud to become visible during the rendering of sparse point clouds, which poses challenges for the image restoration model in discerning the front-to-back relationship and consequently impacts the performance of the restoration network. Conversely, a larger PS can enhance the distinction between the foreground and background of the point cloud, but it also increases the likelihood of points overlapping, leading to a loss of point information. Therefore, we have rendered the point cloud with two different point sizes to address these concerns during restoration and set the $PS$ to 1 and the upsampling factor of $R$, respectively.

\emph{Dual Scale Rendered Image Restoration Network.}
After obtaining the rendered images of the sparse point cloud, we employ an image restoration network to restore the sparse point cloud renderings, with the goal of achieving rendering results comparable to those of a dense point cloud. 
With lightweight considerations in mind, we have chosen BRNet~\cite{yang2021ntire} as our backbone for image restoration. BRNet enhances FFDNet~\cite{zhang2018ffdnet} by incorporating a mean shift module to normalize the input image, thereby achieving better performance. To combine the advantages and mitigate the shortcomings of point cloud renderings with large and small $PS$, we expanded the input channels of FFDNet to six channels and obtained a dual scale restoration network. By concatenating the two point cloud renderings of different point sizes, we input them together into the point cloud restoration network to obtain a dense point cloud rendering with a small size. For more details on the network architecture, please refer to BRNet~\cite{yang2021ntire}.
During network training, considering that there are certain areas of blank background in the point cloud rendering, we assign different loss weights to the foreground and background regions of the point cloud by using the following loss function:
\begin{equation}
\mathcal{L}_{Res} = \alpha \cdot \frac{1}{N_w} \sum{i \in \mathcal{W}} |I_h - \hat{I_h}| + \beta \cdot \frac{1}{N_o} \sum{j \in \mathcal{O}} |I_h - \hat{I}_h|
\end{equation}
where $I_h$ and $\hat{I}_h$ denote the restored rendered image and its ground truth, $\mathcal{W}$ represents the set of white pixels (i.e., the background region with pixel values of (1, 1, 1) and $\mathcal{O}$ represents the set of other pixels (i.e., the foreground region). By setting different values for the weight coefficients $\alpha$ and $\beta$, we can adjust the importance of the foreground and background regions in the total loss, allowing the network to focus more on the foreground while reducing the impact of the background on the overall loss.

\subsection{Gaussian Interpolation}
To better control the desired number of points achieved during upsampling, we perform Gaussian interpolation on the input point cloud before feeding the sparse point cloud $Psp$ into the 3DGS upsampling module. To facilitate parallel computation, we employ the reparameterization sampling technique to perform Gaussian interpolation on the point cloud. The specific sampling formula is as follows:
\begin{equation}
f = \mu + \sigma\varepsilon , \varepsilon \sim \mathcal{N}(\varepsilon; 0, 1)
\end{equation}
In the equation, $u$ represents either the geometric or color attributes of the point and $\sigma$ denotes the variance of the Gaussian distribution. $\mathcal{N}$ represents the standard normal distribution. For the point cloud geometry, we set $\sigma$ to 0.25 times the point cloud step size, while for the point cloud color, we set $\sigma$ to 0. By performing R iterations of reparameterization on each attribute of every point in the sparse point cloud, we obtain the Gaussian interpolated point cloud $P_{gi}$ with R times the number of $P_{sp}$.

\subsection{3DGS Point Cloud Upsampling Module}
After obtaining the restored multi-view images, we utilize them for 3DGS upsampling. 
We can obtain the point cloud geometry and color information based on the center $x$ of the 3D Gaussians and the spherical harmonic coefficients $c$ after optimizing 3DGS~\cite{kerbl20233d}.
However, the number of points in the point cloud generated by the original 3DGS is not constrained, making it difficult to meet our requirement of upsampling to a specified number of points. Moreover, the quality of the generated point cloud is suboptimal. To address these issues, we propose a series of improvements to the vanilla 3DGS, tailoring it specifically for the point cloud upsampling task.

Given a set of point cloud-rendered images and the corresponding camera poses, 3DGS~\cite{kerbl20233d} is able to learn a 3D scene representation in the form of a set of 3D Gaussians. This allows for the rendering of a new image from any desired viewpoint. To better adapt 3DGS for the task of colored point cloud upsampling and achieve improved perceptual quality, we propose the following modifications to the vanilla 3DGS framework:

To precisely achieve the expected number of points after $R\times$ upsampling, we disable the cloning, splitting, and pruning operations in the vanilla 3DGS during the optimization process. This fixes the number of 3D Gaussians, ensuring that the number of points in the point cloud remains consistent with the input $P_{gi}$ fed into the 3DGS upsampling module.

Leveraging the prior knowledge that each point in the point cloud is rendered with the same size, we introduce a scale constraint to ensure consistency across the entire point cloud. Specifically, at each optimization iteration, we compute the mean scale value across all points in the point cloud, considering their three-dimensional scale components. This mean scale value is then assigned to all points, guaranteeing that they share a uniform scale during rendering. Based on the upsampling task setting in this paper, which considers only the RGB channels as color information, we set the opacity of the 3D Gaussian to 1 and disabled the optimization of opacity.

In addition to the L1 distance and DS-SSIM of rendered images used as optimization objectives in the vanilla 3DGS, we also introduce regularization constraints on the upsampled point cloud. Specifically, we use the L1 distance between the color information of the Gaussian interpolated point cloud $P_{gi}$ as the color constraint term. Moreover, we employ the Chamfer Distance (CD) between the geometric information of the original sparse point cloud $P_{sp}$ as the geometric constraint term.
During upsampling, we incorporate the Chamfer Distance every 256 steps within the initial $M$ steps to supervise the geometry. After $M$ steps, we disable the optimization of geometric information and only optimize the color information of the point cloud. Our optimization objective for upsampling using 3DGS is as follows:

\begin{equation}
\footnotesize
\mathcal{L}_{Up} =
\begin{cases}
\lambda_1\mathcal{L}_1(I_\mathrm{r},I_\mathrm{h}) + \lambda_1\mathcal{L}_1(P_\mathrm{gi}^c,P_\mathrm{ds}^c) 
 + \lambda_2\mathcal{L}_\mathrm{D-SSIM}(I_\mathrm{r},I_\mathrm{h}), \\[0.5em]
\quad \text{if } \mathrm{step} \bmod 256 \neq 0 \text{ or } \mathrm{step} > M, \\[1em]
\lambda_1\mathcal{L}_1(I_\mathrm{r},I_\mathrm{h}) + \lambda_1\mathcal{L}_1(P_\mathrm{gi}^c,P_\mathrm{ds}^c) + \lambda_2\mathcal{L}_\mathrm{D-SSIM}(I_\mathrm{r},I_\mathrm{h}) \\
\quad + \lambda_3\mathcal{L}_\mathrm{CD}(P_{\mathrm{sp}}^g,P_{\mathrm{ds}}^g), \\[0.5em]
\quad \text{otherwise},
\end{cases}
\end{equation}
where $\lambda$ are weighting factors. $P^c$ and $P^g$ respectively represent the color and geometric attributes of the point cloud $P$. By optimizing $\mathcal{L}_{Up}$, we are able to achieve a more precise upsampled point cloud.

\begin{figure}
	\centering
	\includegraphics[width=1.0\columnwidth]{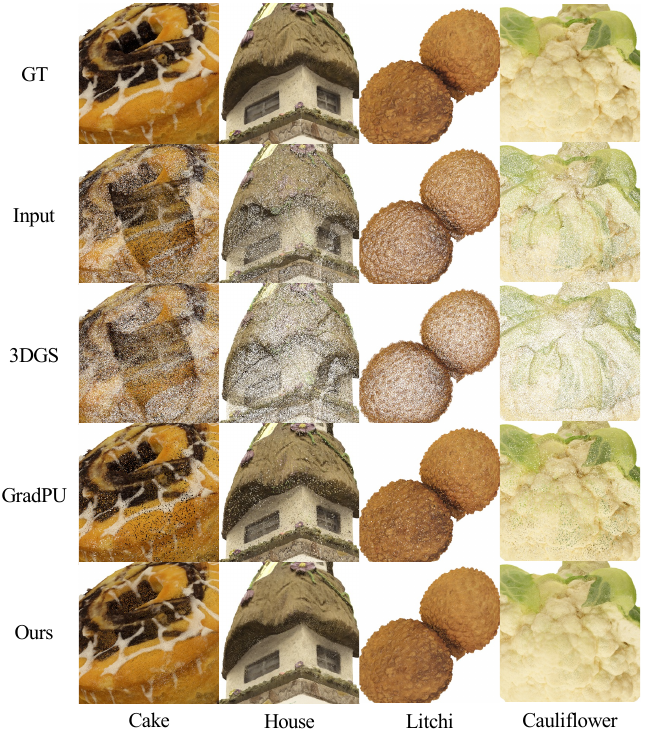} 
	\caption{Visualization of $4\times$ upsampling results. Our method achieves the best quality, exhibiting uniform and smooth surfaces with finer-grained detail representation. The upsampled point clouds show significant improvements in quality compared to the input sparse point clouds.
	}
	\label{figure_compare_render}
\end{figure}

\section{EXPERIMENTS}

\subsection{Experimental Settings}
\emph{Dataset.}
To evaluate the efficacy of our proposed method for color point cloud upsampling, we curated a dataset consisting of 25 dense 10-bit color point clouds from the WPC dataset~\cite{su2019perceptual,liu2022perceptual}, with point counts spanning from 412,009 to 2,878,318, to serve as ground truth for the upsampling task. Among them, five point clouds, each with over 1 million points, were used for testing. The test set includes "Cake" with 2,486,566 points, "Cauliflower" with 1,936,627 points, "House" with 1,568,490 points, "Litchi" with 1,039,942 points, and "Pumpkin" with 1,340,343 points. We performed random 4× downsampling on these dense point clouds using Monte Carlo sampling, reducing the number of points to 1/4 of the original count to obtain sparse point clouds.

\emph{Implementation Details.}
We utilized the Adam optimizer to conduct 100 training epochs for the restoration network. The network's learning rate was configured at 1e-4, with a corresponding weight decay for optimization at 1e-5 As for the parameters within the network's loss function, we designated the values of $\alpha$ and $\beta$ to 0.3 and 0.7, respectively. 
During the Gaussian optimization stage, we carried out a total of 50,000 optimization iterations. In the optimization process, the maximum number of iterations $M$, where the Chamfer Distance is involved, is set to 20,000. We set the weight $\lambda_1$ for the L1 distance to 0.8 while maintaining the weight $\lambda_2$ for D-SSIM at 0.2, consistent with the original 3DGS implementation. The weight $\lambda_3$ for the Chamfer Distance is set to half the number of points in the input point cloud. The degree of the spherical harmonics function is set to 0. All remaining parameters are kept in line with the vanilla 3DGS.
The training and testing of the point cloud rendering restoration network, as well as the optimization process of the 3DGS upsampling module, were executed utilizing the PyTorch~\cite{paszke2019pytorch}. For the experiments, we employed a single NVIDIA GeForce RTX 3090 GPU for computation.

\emph{Evaluation Metrics.}
To quantitatively assess the perceptual quality of color point cloud upsampling, we rendered images from 32 randomly selected viewpoints and computed four key metrics: PSNR, SSIM~\cite{wang2004image}, LPIPS~\cite{zhang2018unreasonable}, and IW-SSIM~\cite{wang2010information}. To minimize the influence of the image background on the PSNR calculation, we focused exclusively on the foreground regions of the rendered images. According to Liu \etal~\cite{liu2022perceptual}, the IW-SSIM metric aligns more closely with the human subjective perception of point cloud quality. Therefore, we recommend IW-SSIM as the primary reference metric for evaluating the perceptual quality of point clouds. Additionally, to quantitatively measure the geometric quality of the upsampled point clouds, we utilized two widely accepted geometric metrics: Chamfer Distance (CD) and Hausdorff Distance (HD).

\begin{table}[]
    \small
    \centering
    \setlength\tabcolsep{5pt} 
    \caption{$4\times$-upsampling quantitative comparison on the WPC dataset. "$\uparrow$" / "$\downarrow$ "indicates that larger/smaller is better. The best results are highlighted in \textcolor{red}{red}.}
    \resizebox{\linewidth}{!}{%
        \begin{tabular}{l|cccccc}
        
            \toprule
            Method     &PSNR $\uparrow$      &SSIM $\uparrow$  &LPIPS $\downarrow$  &IW-SSIM $\uparrow$  & \begin{tabular}[c]{@{}c@{}}CD $\downarrow$\\  10\textasciicircum{}5\end{tabular} & \begin{tabular}[c]{@{}c@{}}HD $\downarrow$ \\ 10\textasciicircum{}2\end{tabular} \\     
            \midrule
            Input                & 13.121               & 0.864                & 0.163                & 0.548                & 0.188                    & 0.771                  \\
3DGS~\cite{kerbl20233d}                & 10.703               & 0.845                & 0.202                & 0.434                & 8.464                    & 7.645                  \\
GradPU~\cite{he2023grad}               & 20.708               & 0.913                & 0.088                & 0.790                & 0.125                    & 0.753                  \\
Ours             & \textcolor{red}{23.684}               & \textcolor{red}{0.932}                & \textcolor{red}{0.043}                & \textcolor{red}{0.876}                & \textcolor{red}{0.124}                    & \textcolor{red}{0.701}                  \\
            \bottomrule
        \end{tabular}%
    }
    \label{table_compare_performance}
\end{table}

\subsection{Performance Comparison}
To the best of our knowledge, the source codes for the advanced counterparts, CloudUP~\cite{cho2023cloudup} and FGTV~\cite{dinesh2020super}, are not publicly available. This unavailability limits our ability to conduct direct comparisons with these models. In this study, we conducted comparative evaluations with our baseline model, the vanilla 3DGS, as well as with the state-of-the-art point cloud geometric upsampling model GradPU~\cite{he2023grad}. For GradPU, we have employed a nearest-neighbor interpolation approach to endow it with color information.

The quantitative experimental comparison results are presented in Table \ref{table_compare_performance}. As shown, our method outperforms both 3DGS and GradPU on perceptual quality assessment metrics as well as geometric measurement indices. Relative to the input sparse point cloud, our approach achieves a 59\% improvement in performance on the IW-SSIM metric. 

To qualitatively compare the results of different upsampling methods, we visualized the ground truth (GT) point clouds, the input sparse point cloud, and the upsampling outcomes from various methods in Figure \ref{figure_compare_render}. Through visualization, it is evident that the point clouds generated by 3DGS are relatively sparse, and the upsampling results from GradPU contain a considerable amount of noise. Our upsampling results surpass those of both 3DGS and GradPU and demonstrate a significant improvement in quality compared to the input sparse point clouds.

\begin{figure}
	\centering
	\includegraphics[width=1.0\columnwidth]{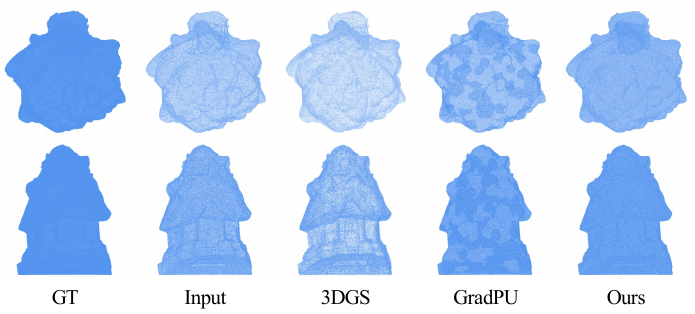} 
	\caption{Geometric visualization of $4\times$ upsampling results. It is evident that our proposed method is capable of generating denser point clouds with higher uniformity and precision compared to other approaches.
	}
	\label{figure_compare_geometry}
\end{figure}

Furthermore, we provide geometric visualizations of the upsampled point clouds, as shown in Figure \ref{figure_compare_geometry}. It can be observed that the point cloud generated by Vanilla 3DGS is relatively sparse. GradPU is affected by the partitioning process, resulting in a noticeable disconnection between the generated point cloud blocks. In contrast, our upsampling approach yields a more uniform, dense, and accurate point cloud. Through the aforementioned quantitative and qualitative comparisons, it is evident that our proposed method demonstrates effective performance and excellent results.

\begin{table}[]
    \small
    \centering
    \setlength\tabcolsep{6pt} 
    \caption{Ablation results on Single Scale and Dual Scale point cloud rendering restoration model.}
    \resizebox{\linewidth}{!}{%
        \begin{tabular}{l|cccc}
        
            \toprule
            Method & PSNR $\uparrow$ & SSIM $\uparrow$ & LPIPS $\downarrow$ & IW-SSIM $\uparrow$ \\     
            \midrule
           Single Scale (PS=1) & 23.765 & 0.939 & 0.070 & 0.807 \\
Single Scale (PS=4)  & 26.556 & 0.950 & 0.061 & 0.882 \\
Dual Scale (Ours) & \textcolor{red}{28.432} & \textcolor{red}{0.961}& \textcolor{red}{0.052} & \textcolor{red}{0.926} \\
            \bottomrule
        \end{tabular}%
    }
    \label{table_ablation_restoration}
\end{table}

\subsection{Ablation Study}
\emph{Dual Scale Restoration Model.}
To validate the effectiveness of our proposed dual-scale point cloud rendering image restoration method, we conducted experiments on both single scale and dual scale restoration models. For the single-scale rendered images, we used rendered point size (PS) as 1 and 4 respectively. The dual scale restoration model takes both two scale rendered images as input simultaneously while keeping other settings consistent. We trained the networks for 100 epochs and used the results from the last epoch for testing.

The quantitative and qualitative experimental results are presented in Table \ref{table_ablation_restoration}. Our proposed dual-scale restoration method demonstrates notable performance improvements compared to the single-scale baseline. Specifically, the PSNR increases by 1.876 dB, and the IW-SSIM improves by 0.044. 

\begin{table}[]
    \small
    \centering
    \setlength\tabcolsep{5pt} 
    \caption{Ablation results on different 3DGS upsampling module configurations.}
    \resizebox{\linewidth}{!}{%
        \begin{tabular}{l|cccccc}
        
            \toprule
            Method     &PSNR $\uparrow$      &SSIM $\uparrow$  &LPIPS $\downarrow$  &IW-SSIM $\uparrow$  & \begin{tabular}[c]{@{}c@{}}CD $\downarrow$\\  10\textasciicircum{}5\end{tabular} & \begin{tabular}[c]{@{}c@{}}HD $\downarrow$ \\ 10\textasciicircum{}2\end{tabular} \\     
            \midrule
           3DGS (Base)                                                     & 10.703 & 0.845 & 0.202 & 0.434   & 8.464                                                               & 7.645                                                               \\
Base+FN                                                           & 14.404  & 0.856 & 0.166 & 0.474   & 0.172                                                               & 1.11                                                               \\
Base+FN+DO                                                        & 14.872
& 0.870 & 0.150 & 0.605   & 0.181                                                               & 0.753                                                              \\
Above+US (Ours) & \textcolor{red}{23.681} & \textcolor{red}{0.932} & \textcolor{red}{0.043} & \textcolor{red}{0.876}   & \textcolor{red}{0.124}                                                               & \textcolor{red}{0.701}   \\ 
            \bottomrule
        \end{tabular}%
    }
    \label{table_ablation_3dgs}
\end{table}

\emph{3DGS Modification.}
To further validate the improvements we made to 3DGS for point cloud upsampling, we conducted a series of ablation experiments. The following configurations were tested: (1) our baseline vanilla 3DGS (Base), (2) Base with Gaussian Interpolation and a fixed number of points (Base+FN),  (3) Base+FN+US with disabled opacity (Base+FN+DO), and (4) Base+FN+DO  (Above) with a uniform 3D Gaussian scale (Above US)

The quantitative results are presented in Table \ref{table_ablation_3dgs}. The improvements in PSNR, SSIM, LPIPS, and IW-SSIM metrics demonstrate that our proposed modifications, including Gaussian interpolation with a fixed number of points, uniform Gaussian scale, and disabling opacity optimization, effectively enhance the perceptual quality of the point cloud. When all three modifications are applied, the IW-SSIM metric reaches its highest value (0.900), representing a 107\% performance increase compared to the baseline (0.434). 
However, these three improvements can not enhance geometric quality. Adding point cloud normalization slightly decreases perceptual quality but significantly improves geometric quality. It results in order-of-magnitude improvements in CD and HD metrics. We deem this trade-off worthwhile. These comparisons demonstrate that our 3DGS modifications can substantially improve performance.

\section{CONCLUSION}
In this paper, we introduced GaussianPU, a novel 2D-3D hybrid colored point cloud upsampling framework that combines 3D Gaussian Splatting (3DGS) and 2D rendered image restoration network.  
Our framework enables the direct upsampling of large-scale colored point clouds on consumer-grade GPUs without point cloud segmentation, thus avoiding the quality degradation that can result from patch-based processing.  
Moreover, we introduced a dual-scale point cloud rendering image restoration network and implemented a series of enhancements to 3DGS tailored for point cloud upsampling.  
These improvements enable precise control over the number of upsampled points and enhance the quality of the upsampled point clouds.  
Experiments show our method greatly improves perceptual and geometric quality of sparse point clouds.  Its ability to process millions of points directly indicates high potential for large-scale robotic applications like autonomous navigation and environmental mapping.

\end{document}